\def\year{2021}\relax
\documentclass[letterpaper]{article} 
\usepackage{aaai21}  
\usepackage{times}  
\usepackage{helvet} 
\usepackage{courier}  
\usepackage[hyphens]{url}  
\usepackage{graphicx} 
\usepackage{natbib}  
\usepackage{caption} 
\usepackage{multirow}
\usepackage[utf8]{inputenc} 
\usepackage[T1]{fontenc}    
\usepackage{hyperref}       
\usepackage{url}            
\usepackage{booktabs}       
\usepackage{amsfonts}       
\usepackage{nicefrac}       
\usepackage{microtype}      
\usepackage[noend]{algorithmic}
\usepackage{algorithm}

\usepackage{amssymb}
\usepackage{amsthm}
\usepackage{graphicx}
\usepackage{subfig}
\usepackage{wrapfig}
\usepackage{color}

\usepackage{amsmath}

\DeclareMathOperator*{\argmax}{arg\,max}

\title{Deep Learning for Earth Image Segmentation based on Imperfect Polyline Labels with Annotation Errors}
\author{
    Zhe Jiang, Marcus Stephen Kirby, Wenchong He, and Arpan Man Sainju\\
}
\affiliations{
Department of Computer Science, The University of Alabama\\
zjiang@cs.ua.edu, \{mskirby1,whe11,asainju\}@crimson.ua.edu

}

\begin{document}

\maketitle

\begin{abstract}
In recent years, deep learning techniques (e.g., U-Net, DeepLab) have achieved tremendous success in image segmentation. The performance of these models heavily relies on high-quality ground truth segment labels. Unfortunately, in many real-world problems, ground truth segment labels often have geometric annotation errors due to manual annotation mistakes, GPS errors, or visually interpreting background imagery at a coarse resolution. Such location errors will significantly impact the training performance of existing deep learning algorithms. Existing research on label errors either models ground truth errors in label semantics (assuming label locations to be correct) or models label location errors with simple square patch shifting. These methods cannot fully incorporate the geometric properties of label location errors. To fill the gap, this paper proposes a generic learning framework based on the EM algorithm to update deep learning model parameters and infer hidden true label locations simultaneously. Evaluations on a real-world hydrological dataset in the streamline refinement application show that the proposed framework outperforms baseline methods in classification accuracy (reducing the number of false positives by {\bf 67\%} and reducing the number of false negatives by {\bf 55\%}).
\end{abstract}

\section{Introduction}\label{sec:intro}
In recent years, deep learning techniques (e.g., U-Net~\cite{ronneberger2015u}, DeepLab~\cite{chen2017deeplab}, SegNet~\cite{badrinarayanan2017segnet}) have achieved tremendous success in image segmentation~\cite{hesamian2019deep,minaee2020image,garcia2018survey,zhou2019review}. The current state of the art approaches consist of a downsample path (through convolutional and max-pooling layers) and an upsample path (through deconvolutional layers together and skip connections). The upsample path combines both global and local semantic features to provide detailed segmentation. 

The performance of deep learning models heavily relies on high-quality ground truth segment labels. Unfortunately, in many real-world problems of high-resolution earth imagery segmentation, such as land cover mapping, ground truth segment labels often have geometric annotation errors~\cite{tajbakhsh2020embracing,jiang2017spatial}. These errors can be due to manual annotation mistakes, particularly if the annotators are non-expert and thus unable to fully interpret some images pixels~\cite{goodchild2012assuring,senaratne2017review,degrossi2018taxonomy,dorn2015quality,herfort2019mapping,jiang2018survey,shekhar2015spatiotemporal}. Annotation errors can also come from GPS errors when a field crew travels on the ground to delineate the boundary of a land parcel. In addition, label errors can be due to annotators visually interpreting background imagery displayed at a coarse resolution (e.g., on a small screen of a smartphone). Such geometric annotation errors can significantly impact the effectiveness of existing deep learning algorithms.

Training deep learning models for earth imagery segmentation based on imperfect labels is non-trivial for several reasons. First, the annotation errors in labels may follow certain geometric properties (e.g., distance and angle). Second, the problem requires a learning algorithm to infer true label locations and train neural network parameters simultaneously. Finally, the problem is computationally challenging due to a large number of potential truth label locations. 

Existing research that addresses label errors often focuses on addressing errors in label semantics, assuming label locations to be correct. Techniques include simple data cleaning to filter noise~\cite{frenay2014classification}, choosing relatively noise-tolerant models~\cite{dietterich2000experimental,abellan2012bagging}, designing robust loss function~\cite{patrini2017making,mnih2012learning,reed2014training} and learning noise distribution~\cite{kajino2012learning,xiao2015learning,Lu2017,rolnick2017deep,xiao2015learning}. Thus, these techniques cannot address the location errors in the ground truth labels. The closest related works that focus on label location errors are~\cite{mnih2012learning,chen2019autocorrect}. These works model location errors of training labels as small shifts of square image patches in eight-neighbor directions. In reality, however, label location errors are often represented as the shifts of vertices in geometric shapes of class labels, which cannot be accurately modeled by shifts of square image patches. 
Other works rely on interactive active learning to address imperfect labels~\cite{yang2017suggestive}, but this approach requires human experts in the loop.

In contrast, this paper proposes a novel location error model for class labels represented by geometric shapes (e.g., spatial points, polylines). Specifically, we propose a geometric error model for the conditional probability of observed (erroneous) point label locations given (an unknown) true label location in the polar coordinate system. We then generalize the error model from points to polylines. Based on the location error model, we propose a generic learning framework through the EM algorithm to jointly update deep learning model parameters and infer hidden true label locations. Evaluations on real-world earth imagery datasets for streamline refinement and road mapping applications show that the proposed framework significantly outperforms baseline methods in classification accuracy.

\section{Problem Statement}\label{sec:prob}
\subsection{Preliminaries}
We denote the features of all pixels each image as $\mathbf{X}\in \mathbb{R}^{n\times n}$, and the corresponding pixel class labels as $\mathbf{Y}\in\{0,1\}^{n\times n}$. Note that both $\mathbf{X}$ and $\mathbf{Y}$ are in grid representation. We denote a true label spatial point location as $\mathbf{l}\in\mathbb{R}^{2\times1}$ and a true polyline label location as a sequence of point locations $\mathbf{L}=<\mathbf{l}_1,\mathbf{l}_2,...,\mathbf{l}_{n_\mathbf{L}}>$. 
We also denote the observed noisy locations of a geometric shape of point as $\tilde{\mathbf{l}}$ and a polyline as $\tilde{\mathbf{L}}=<\tilde{\mathbf{l}}_1,\tilde{\mathbf{l}}_2,...,\tilde{\mathbf{l}}_{ n_{ \tilde{\mathbf{L}}} }>$. Note that the geometric representation of training labels (i.e., $\mathbf{L}$ and $\tilde{\mathbf{L}}$) and the grid representation (i.e., $\mathbf{Y}$ and $\tilde{\mathbf{Y}}$) are equivalent and mutually exchangeable through rasterization or vecterization operations. We denote the deep learning model for image segmentation as $\mathbf{Y}=f(\mathbf{X};\mathbf{\Theta})$, where $\mathbf{\Theta}$ is neural network parameters.


\subsection{Problem definition}
Given a base deep learning model with parameters $\mathbf{\Theta}$, feature images $\mathbf{X}$, and ground truth segment labels with geometric annotation errors in the form of polylines $\tilde{\mathbf{L}}$, the problem aims to learn the model parameters $\mathbf{\Theta}$.

\begin{figure}
    \centering
    \includegraphics[width=2in]{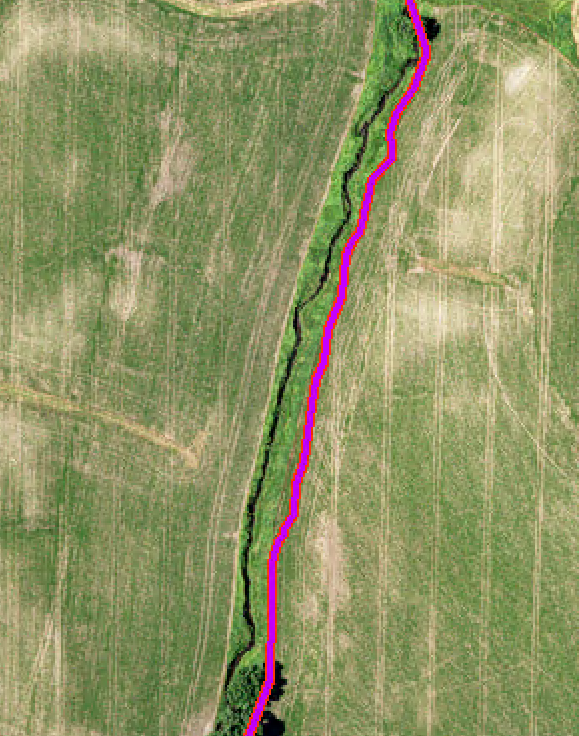}
    \caption{A real-world problem example in streamline segmentation from high-resolution earth imagery. The purple line is an imperfect label that is misaligned with the true streamline location.}
    \label{fig:probeg}
\end{figure}

Figure~\ref{fig:probeg} provides a real-world example in streamline segmentation from high-resolution earth imagery. The input ground truth training labels (in purple) are misaligned with the true stream locations in the earth imagery (in black color). Thus, directly training a deep learning model from the imperfect label will lead to poor classification performance. Given geometric shapes of class labels with registration error (misaligned with image pixels), earth imagery with explanatory features (e.g., spectral bands), as well as a base deep learning model (e.g., U-Net~\cite{ronneberger2015u}, DeepLab~\cite{chen2017deeplab}), the problem aims to find a robust learning algorithm that can train an accurate deep learning model. 

\section{The Proposed Approach}\label{sec:app}
The goal is to investigate novel spatial deep learning algorithms that are robust to location registration errors in geometric shapes of class labels. Geometric shapes (e.g., point of interest, road lines, and building polygons) from non-expert volunteers often contain location errors due to manual annotation mistakes, GPS errors, or interpreting coarse-resolution imagery. The problem is challenging due to the uncertainty of true geometric shape locations and the high computational costs associated with learning from uncertain ground truth locations. Most deep learning research on label errors focuses on label semantic noise~\cite{sukhbaatar2014learning,reed2014training,patrini2017making} instead of location errors. Existing works on label location error often model ground truth classes as square image patches and fail to consider geometric properties~\cite{mnih2012learning,chen2019autocorrect}.
In contrast, we propose to investigate a new location error model that better captures geometric properties of label shapes and efficient algorithms in a joint learning framework to update deep learning model parameters based on uncertain true shape locations. 

\begin{figure*}
    \centering
	\includegraphics[width=4in]{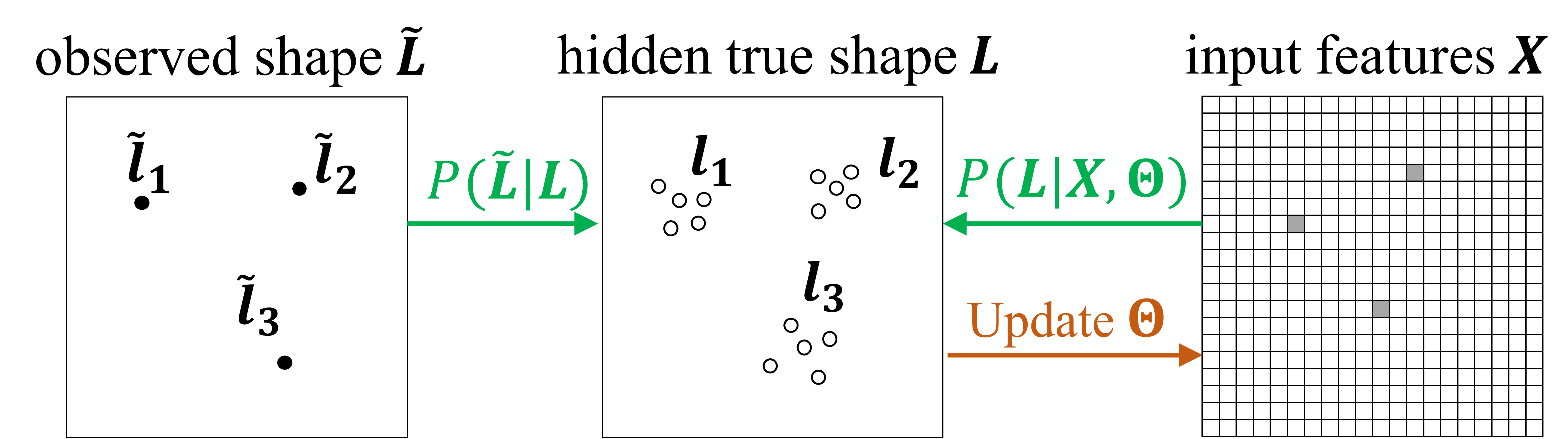}
    \caption{Illustration of the EM framework for point labels (green arrow shows the E-step, brown arrow shows the M-step)}
    \label{fig:task2b}
\end{figure*}
\subsection{Statistical model of point location registration error}
This step aims to explore statistical models for geometric errors between observed shapes and true class shapes. For simplicity, we start with spatial point labels. More complex shapes such as polylines and polygons can be considered collections of points and will be considered later. Given an observed noisy two-dimensional point location $\tilde{\mathbf{l}}=(\tilde{l}^{(1)},\tilde{l}^{(2)})$ and its true location $\mathbf{l}=(l^{(1)},l^{(2)})$, this step aims to find a statistical model for $P(\tilde{\mathbf{l}}|\mathbf{l})$. The task is non-trivial for two reasons. First, a geometric error can have an arbitrary distance and angle. Second, there can be a very large (potentially infinite) number of true shape locations corresponding to each observed geometric shape.

\begin{figure}
    \centering
    \subfloat[Point location error]{\includegraphics[width=1in]{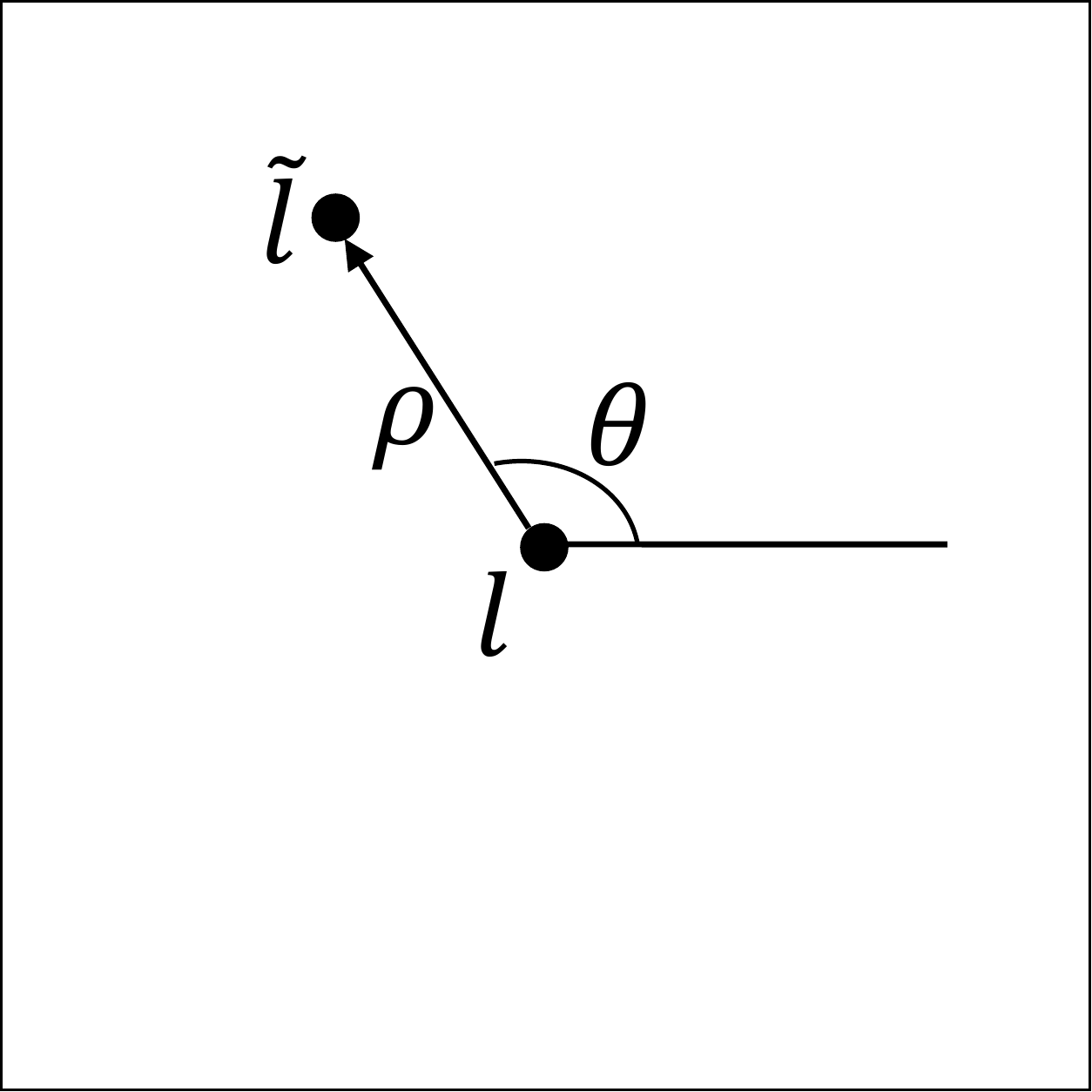}}
    \hspace{1mm}
    \subfloat[Discretized point location error]{\includegraphics[width=1in]{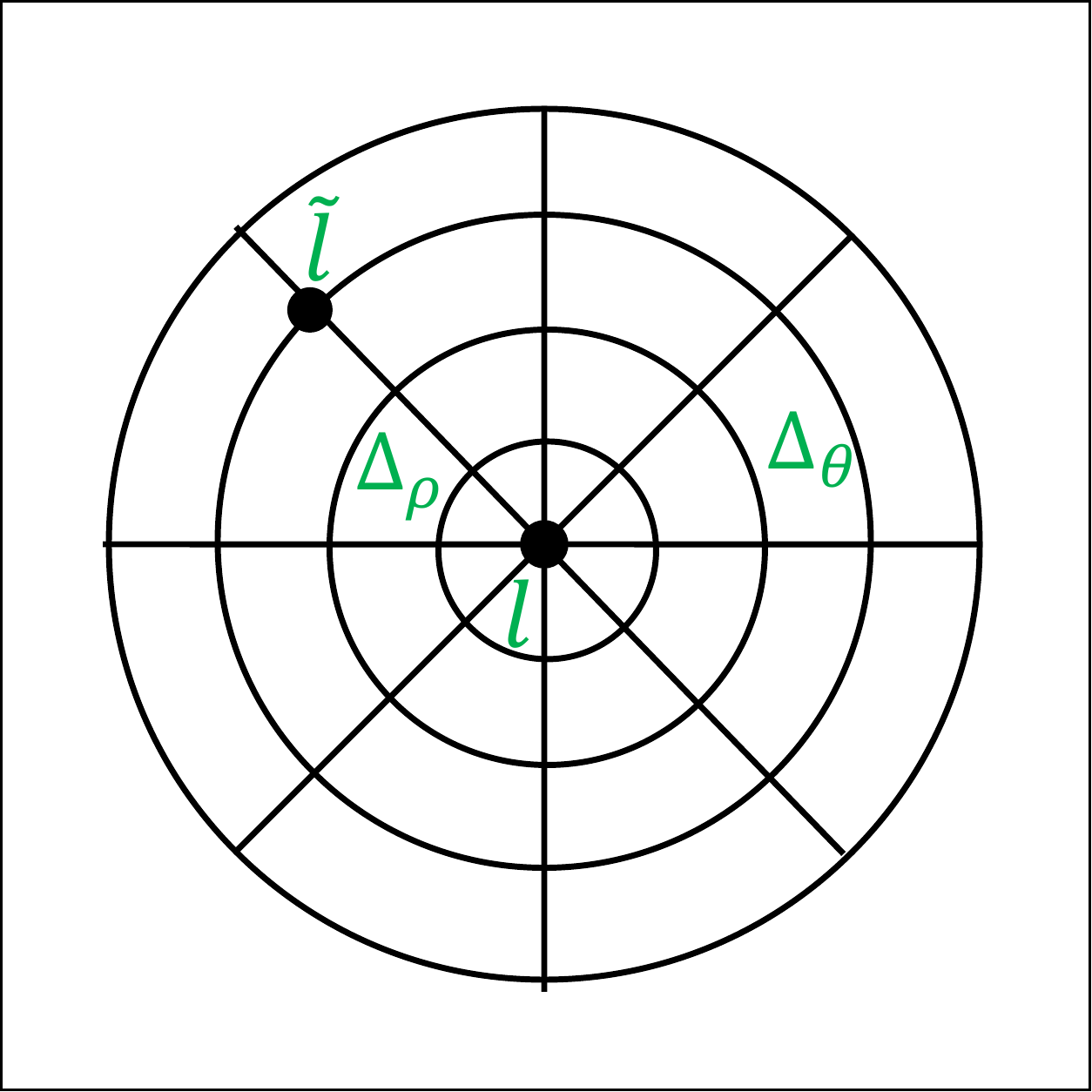}}
    \caption{Statistical model for point location error in a polar coordinate system}
    \label{fig:task2a}
\end{figure}

To address the challenge of an arbitrary distance and angle, we propose a location error model within a polar coordinate system. Figure~\ref{fig:task2a}(a) provides an example. The difference between an erroneous observed point location and the underlying true location can be expressed by $(\rho,\theta)$, where $\rho$ is the distance and $\theta$ is the angle. Assuming the observed location $\tilde{\mathbf{l}}$ is known and fixed, there is a 
one-to-one mapping between any particular value of $(\rho,\theta)$ and an underlying true location $\mathbf{l}$ through the equation $(\tilde{l}^{(1)},\tilde{l}^{(2)})=(l^{(1)}+\rho\cdot\cos(\theta) ,l^{(2)}+\rho\cdot\sin(\theta))$. In this way, location uncertainty $P(\tilde{\mathbf{l}}|\mathbf{l})$ can be captured by the distribution of $(\rho,\theta)$. We can assume that the error angle $\theta$ follows a uniform distribution within the interval of $[0,2\pi]$, and error distance $\rho$ follows a uniform distribution $[0, \rho_{max}]$. Distribution parameters such as $\rho_{max}$ can be considered as hyper-parameters that could be chosen by a validation data or domain knowledge on the maximum spatial scale of location errors. To address this challenge of infinite number of potential true point locations, we plan to investigate the discretization of location errors $(\rho,\theta)$ into a finite number of values. Specifically, $(\rho,\theta)=(k_{\rho}\Delta_\rho,k_{\theta}\Delta_\theta)$, where $\Delta_\rho$ and $\Delta_\theta$ are hyper-parameters that controls the error resolution, and $k_{\rho}$ and $k_{\theta}$ are random variables that control the magnitude of error distance and error angle. Figure~\ref{fig:task2a}(b) provides an illustrative example, where $\Delta_\theta=\pi/4$, $k_{\rho}=3$ and $k_{\theta}=3$. An additional challenge is that the ground truth point shapes are often in the form of multi-point (a collection of points) instead of a single point. In this case, we can assume different points in the set are independent since they belong to different geometric shapes. We denote $\tilde{\mathbf{L}}=\{\tilde{\mathbf{l}}_i|1\leq i \leq n\}$ and ${\mathbf{L}}=\{{\mathbf{l}}_i|1\leq i \leq n\}$ as the set of observed noisy locations and the  set of true locations respectively, where $n$ is the number of points in the set. Assuming that different points are independent from each other, we have $P(\tilde{\mathbf{L}}|\mathbf{L})=\prod_{i=1}^n P(\tilde{\mathbf{l}_i}|\mathbf{l}_i)$. 

\subsection{Joint learning via EM}
This step aims to explore effective and efficient algorithms to learn deep models from class labels with location errors. We aim to develop a general learning framework without assuming any particular deep learning model (e.g., U-Net~\cite{ronneberger2015u}, DeepLab~\cite{chen2017deeplab}). The task is very challenging for two reasons. First, both the true locations of label points and the deep learning model parameters are unknown, while existing deep learning algorithms often assume that training data are known. Second, the problem is also computationally challenging due to the high computational overhead associated with many possible true class point locations. 

We propose a deep learning framework based on the expectation-maximization (E-M) algorithm~\cite{bishop2006pattern} to address the first challenge. The E-M algorithm is a common strategy in machine learning for jointly estimating both latent variables (e.g., true label locations) and unknown parameters (e.g., deep models). The main idea is 
illustrated in Figure~\ref{fig:task2b}. The framework first initializes deep learning model parameters $\mathbf{\Theta_0}$, e.g., through pre-training a model from observed label points with errors $\mathbf{\tilde{L}}$. Then, it estimates the posterior distribution of true label point locations $P(\mathbf{L}|\mathbf{\tilde{L}},\mathbf{X},\mathbf{\Theta_0})$ based on location error model $P(\mathbf{\tilde{L}}|\mathbf{L})$ from Task 2A and the deep model response $P(\mathbf{L}|\mathbf{X},\mathbf{\Theta_0})$. As the figure shows, there can be multiple true point locations corresponding to one observed point (imperfect label location), each associated with a probability. Next, based on the probabilities of true locations, it can compute the expected loss of the current deep learning model and use that to update deep learning model parameters with gradient descent. In other words, instead of updating model parameters with loss function over a known ground truth location, the framework uses expected loss function over multiple potential ground truth locations (weighted by their posterior probabilities). 

The iterations will continue until stop criteria are met (e.g., after a number of iterations, no updates in parameters).
The specific formulas of the EM iterations are shown in Equation~\ref{eq:estep} and Equation~\ref{eq:mstep}, where $\log{P(\mathbf{L}|\mathbf{X},\mathbf{\Theta})}$ can be considered as the negative of loss function in a base deep learning model. The question now is how to estimate the posterior distribution of true class point locations. We plan to explore potential solutions based on Bayes' theorem. Specifically, $P(\mathbf{L}|\tilde{\mathbf{L}},\mathbf{X},\mathbf{\Theta_0})=\frac{P(\tilde{\mathbf{L}}|\mathbf{L})P(\mathbf{L}|\mathbf{X},\mathbf{\Theta_0})}{\sum_{\mathbf{L}}{P(\tilde{\mathbf{L}}|\mathbf{L})P(\mathbf{L}|\mathbf{X},\mathbf{\Theta_0})}}$, which involves enumerating all potential configuration of true point locations.  In the M-step (Equation~\ref{eq:mstep}), we can potentially leverage the chain-rule to compute gradients over all model parameters. 
\begin{equation}\label{eq:estep}
\begin{split}
\text{E-Step:}~&\mathbb{E}_{\mathbf{L}|\tilde{\mathbf{L}},\mathbf{X},\mathbf{\Theta_0}}\log{P(\mathbf{L}|\mathbf{X},\mathbf{\Theta})}=\\
&\sum_{\mathbf{L}} P(\mathbf{L}|\tilde{\mathbf{L}},\mathbf{X},\mathbf{\Theta_0})\log{P(\mathbf{L}|\mathbf{X},\mathbf{\Theta})}
\end{split}
\end{equation}

\begin{equation}\label{eq:mstep}
\text{M-Step:}~\mathbf{\Theta}\leftarrow \argmax_{\mathbf{\Theta}} \mathbb{E}_{\mathbf{L}|\tilde{\mathbf{L}},\mathbf{X},\mathbf{\Theta_0}}\log{P(\mathbf{L}|\mathbf{X},\mathbf{\Theta})}
\end{equation}

\begin{figure*}
    \centering
	\includegraphics[width=4in]{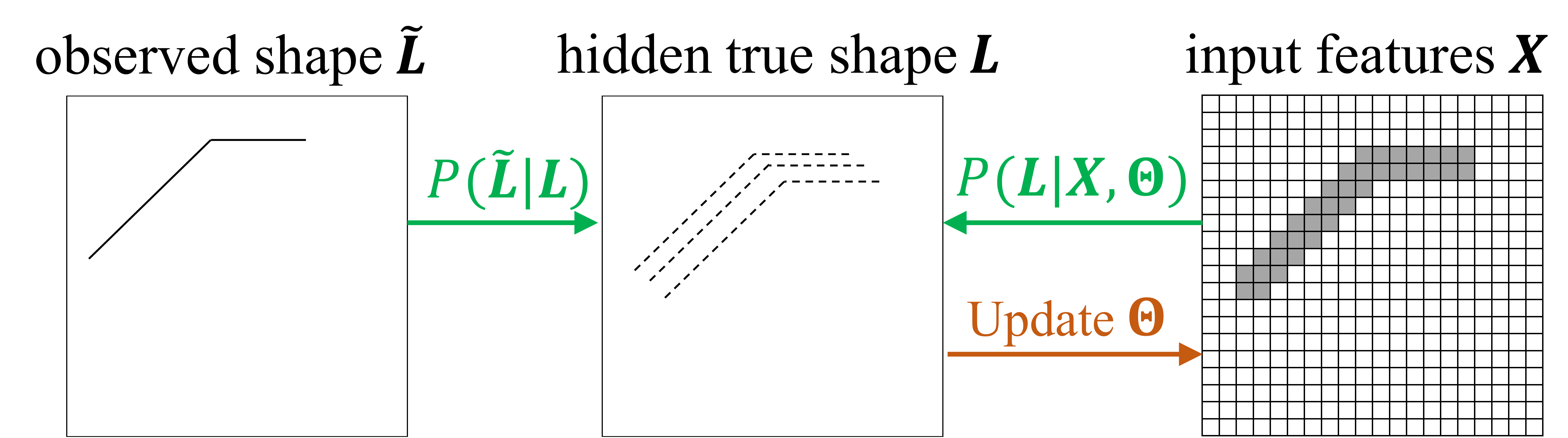}
    \caption{Illustration of the EM framework for polyline labels}
    \label{fig:task2c}
\end{figure*}

We conduct computational bottleneck analysis and investigate potential computational refinements to address the challenge of high computational costs. Initial analysis shows that the main computational risk is that there are too many potential configurations of true point locations, each of which needs to be considered in the expected loss function for parameter learning. The issue is particularly significant when there are multiple observed points due to combinatorics. A brute force method to compute Equation~\ref{eq:estep} is infeasible. To address this issue, we need to explore potential simplifications. For example, we can assume that different label points in the observed set are well separated from each other (their distances are larger than the maximum location error of each point). In this case, we can explore a divide-and-conquer strategy to decompose the loss function over different components; each is associated with one observed point. Specifically, based on the independence assumption between individual points, Equation~\ref{eq:estep} can be simplified into $\sum_i\sum_{\mathbf{l}_i}  P(\mathbf{l}_i|\tilde{\mathbf{l}_i},\mathbf{X},\mathbf{\Theta_0})\log{P(\mathbf{l}_i|\mathbf{X},\mathbf{\Theta})}$, where $i$ is the index of point. Thus, the expected loss over different points can potentially be computed separately.
We need to make a trade-off between model accuracy and computational efficiency when choosing hyper-parameters $\Delta_\rho$ and $\Delta_\theta$. Small values of $\Delta_\rho$ and $\Delta_\theta$ could potentially model location errors more precisely but will also increase computational costs.

\subsection{Generalization from points to polylines}
This step aims to generalize our framework from point class labels to polyline, as illustrated in Figure~\ref{fig:task2c}. The task is non-trivial for two main reasons. First, though a polyline or polygon can be considered as a set of points (vertices), the points (vertices) within a polyline or polygon are not independent. For example, if a vertex is shifting upward from its true pixel location, the next vertex is more likely to shift upward. Second, the complexity of polyline or polygon shapes also significantly increases the computational costs of iterative learning algorithms. 

We propose to generalize our location error models by incorporating the spatial autocorrelation effect between adjacent vertices to address the first challenge. Specifically, the error distance $\rho$ (or $k_{\rho}$) and error angle $\theta$ (or $k_{\theta}$) of two adjacent vertices should be close to each other. For example, we can model the joint distribution of location errors on all vertices within a polyline or polygon as a Markov random field. Without the loss of generalizability, we can assume that the number of vertices in the true polyline (or polygon) and that in the observed polyline (or polygon) are the same. For the case when those numbers are different, we can easily add dummy vertices along a line segment to make the numbers even. 

To address the second challenge, we can conduct computational optimizations on learning algorithms. There are several major computational risks. First, there can be many vertices within an observed polyline or polygon shape, dramatically increasing the enumeration space of true shape locations due to combinatorics. This makes the computation of the E-step (summation over $\mathbf{L}$ in Equation~\ref{eq:estep}) very expensive. To address this risk, we can use computational pruning. For example, we can divide a polyline with too many vertices into smaller polylines and thus reduce the enumeration space of true shape locations. For example, we can divide the polyline in Figure~\ref{fig:task2c} into smaller line segments while maintaining the spatial autocorrelation of location errors at adjacent vertices within a segment. In addition, we can potentially prune out a candidate true shape location if its estimated posterior probability is too small (these locations are unlikely to be the true label locations) and only focus on the top-$K$ candidates ($K$ is a hyper-parameter). Second, evaluating the loss function corresponding to all potential true polyline locations (Equation~\ref{eq:estep}) requires the rasterization of those shapes, which can be very expensive for a large number of potential locations. This computational bottleneck can potentially be addressed by heterogeneous computing to conduct rasterization using multi-thread parallelization on CPU. 
\subsection{Overall algorithm structure}
\begin{algorithm}\small
\caption{The overall learning algorithm}
\label{alg:learning}
\begin{algorithmic}[1]
\REQUIRE\quad\\
$\bullet$ $\mathbf{X}$: {Earth imagery pixel features}\\
$\bullet$ $\tilde{\mathbf{L}}$: {Imperfect ground truth label locations}\\
\ENSURE\quad\\
$\bullet$ $\mathbf{\Theta}$: {Neural network parameters}\\
\STATE Pre-train a neural network $\boldsymbol{\Theta}$ based on imperfect labels $(\mathbf{X},\tilde{\mathbf{L}})$
\STATE Generate the set of candidate true locations $\mathbf{L}$
\WHILE{stop criteria not satisfied}
    \STATE Compute $P(\mathbf{L}|\tilde{\mathbf{L}},\mathbf{X},\mathbf{\Theta_0})$ by model outputs
    \STATE Compute loss function $\mathbb{E}_{\mathbf{L}|\tilde{\mathbf{L}},\mathbf{X},\mathbf{\Theta_0}}\log{P(\mathbf{L}|\mathbf{X},\mathbf{\Theta})}$
    \STATE Update $\boldsymbol{\Theta}$ by the weighted loss function\\
\ENDWHILE
\RETURN $\Theta$
\end{algorithmic}
\end{algorithm}

Algorithm~\ref{alg:learning} provides an overview of the algorithm structure. The algorithm first pre-trains a deep learning segmentation model (e.g., U-Net) from imperfect ground truth labels (after rasterization). The preprocessing also generates a set of candidate true locations of every segment of the imperfect polyline shape. Then, the algorithm goes through EM iterations. In each iteration, the model first calculates a weight for every candidate true locations. The weight is measured by the class likelihood according to deep learning model class probabilities. We can compute the revised loss function with the weighted true segment locations and re-train neural network parameters with gradient descent. 
\section{Evaluation}\label{sec:eval}
\begin{table*}
\centering
\caption{Comparison on classification performance}
\begin{tabular}{|c|c|c|c|c|c|c|}
\hline
Method & Class &\multicolumn{2}{|c|}{Confusion Matrix} &Precision & Recall  &F score\\ \hline
\multirow{2}{*}{U-Net} &Non-stream &9750497&{147480}&0.99&0.99&0.99\\ \cline{2-7}
&{\bf Stream}&{79854}&57369&{\bf 0.39}&{\bf 0.57}&{\bf 0.46}\\ \hline

\multirow{2}{*}{Our Method} &Non-stream&9849319&{48658}&1.00&1.00&{1.00}\\ \cline{2-7}
&{\bf Stream}&{44303}&92920&{\bf 0.66}&{\bf 0.68}&{\bf 0.67}\\ \hline
\end{tabular}
\label{tab:compare}
\end{table*}
The goal is to compare the proposed model with the baseline method in classification performance. We will also analyze the training curves and parameter sensitivity of the proposed model. All experiments were conducted on a deep learning workstation with 4 NVIDIA RTX 6000 GPUs connected by NV-Link (each with 24GB GPU memory) and 128 GB RAM. For the baseline method, We used U-Net~\cite{ronneberger2015u} implemented in Python and Keras (source codes~\cite{u-net}). {\bf Note that more details in experiment setup and additional experiment results were provided in the supplemental materials.}

{\bf Dataset description}: We evaluated our proposed method in a real-world application of refining the National Hydrography Dataset (NHD) based on high-resolution remote sensing data. NHD is a widely used digital database of surface water features such as hydrological streamlines. With the increasingly available high-resolution remote sensing imagery and LiDAR data, the US Geological Survey is currently conducting a refinement of NHD to a higher resolution for the next generation of the hyper-resolution national water model. We used a dataset collected from the Rowan Creek in Wisconsin, USA. The input features include earth imagery from the National Agriculture Imagery Program (NAIP) with red, green, blue, and near-infrared channels, digital elevation model, Lidar point cloud intensity, and slope derived from elevation. The input imperfect streamline location shapefile was collected from an earlier coarse version of NHD. The ground truth streamline locations for testing is manually refined by hydrologists (this true location was hidden from our model in training and validation and was only used for testing). All imagery was resampled into a 1-meter resolution. We used a 2-meter buffer to rasterize training polylines.

We split the area into two halves: the upper half is for testing, and the lower half is for training and validation. In the lower half, we randomly selected 698 windows for training and 40 windows for validation. The training windows and validation windows are not overlapping with each other to preserve independence. In the upper half, we randomly selected 200 windows in the upper half for testing. We augmented training and validation windows by flipping horizontally and vertical as well as 90-degree rotation. Thus, the total number of training and validation windows was 2792 and 160, respectively. The window size is 224 by 224 pixels. Note that in the training and validation windows, we used imperfect lines. We only used manually refined lines in testing windows.

{\bf Model hyper-parameters}:
For the U-Net model, we used double-convolution layers and batch normalization. The dropout rate is $0.2$. We used the negative of dice co-efficient as the loss function. The dice co-efficient loss function is the same as F1-score except that it allows for soft predicted class probabilities. We used a decaying learning rate that reduced the learning rate by half if the validation loss did not improve over 5 epochs (with an initial learning rate of $10^{-1}$ and a minimum learning rate of $10^{-5}$. We also used early stopping to stop model fitting if the validation loss did not improve over 20 epochs. We used a maximum of $50$ epochs in model pre-training and each EM iteration. 

For candidate true shape location generations in our method, we split the input imperfect polylines into small chunks (each with a length of $20$ meters). For simplicity, we generated candidate true locations by shifting the segment in perpendicular directions (9 candidates above and 9 candidates below, 19 in total include the input shape segment itself). Since computing the expected (weighted) loss function in our EM was computationally very expensive due to combinatorics over segments, we simulated the expected loss by the sampling method, i.e., selecting a single candidate for each polyline chunk according to its likelihood probability. We also added a small chance (with $\epsilon$ probability) to randomly select a true location from top-$K$ candidate locations (based on their weights from the current U-Net predicted class probability map). We set $\epsilon=0.05$ and $K=5$. Within each EM iteration, we re-trained our model from scratch with the newly inferred label locations.

{\bf Evaluation metrics}: We used precision, recall, and F-1 score on the streamline class to evaluate candidate methods. 

\begin{figure*}
    \centering
    \subfloat[EM iteration 1]{\includegraphics[width=2.1in]{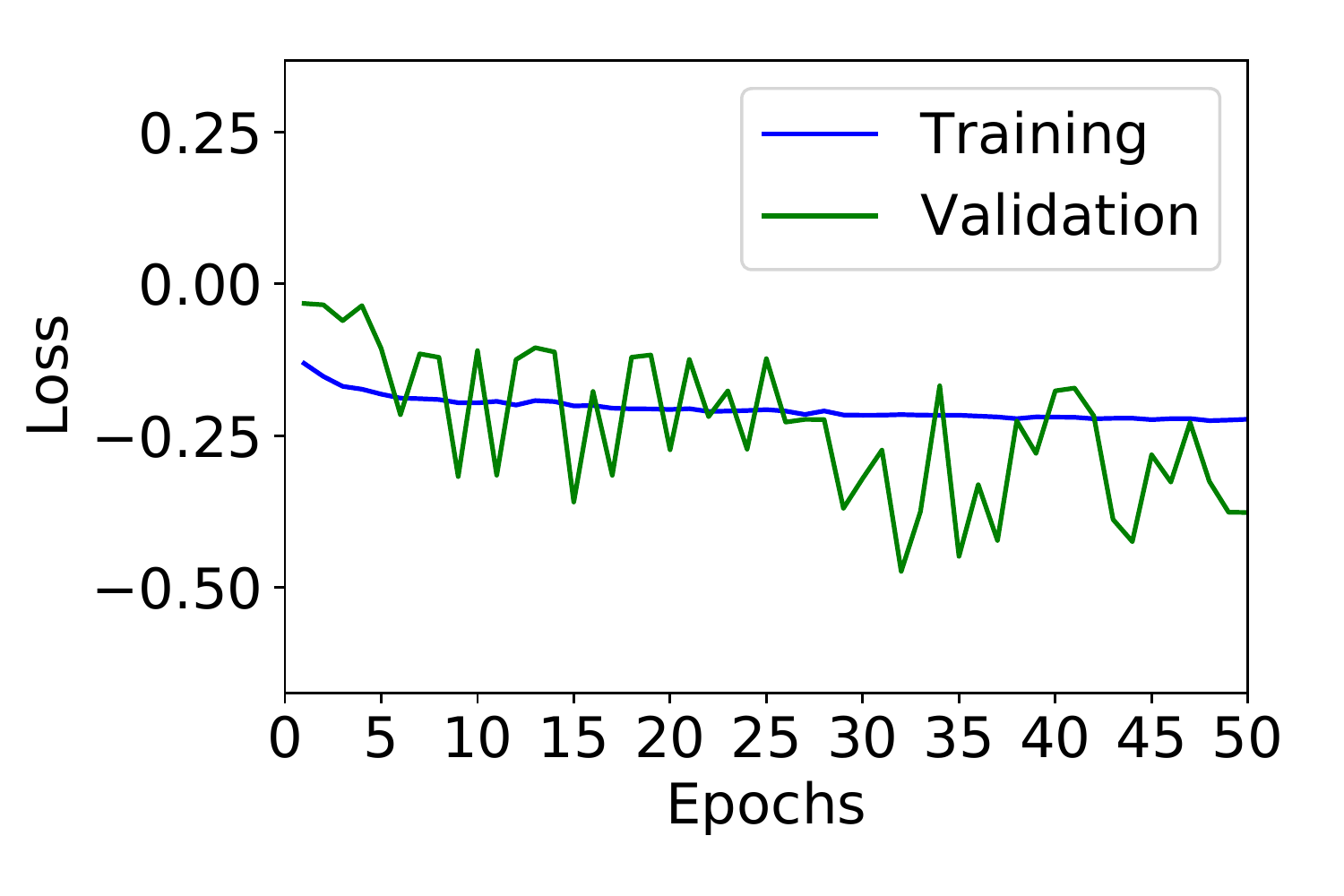}}       \subfloat[EM iteration 2]{\includegraphics[width=2.1in]{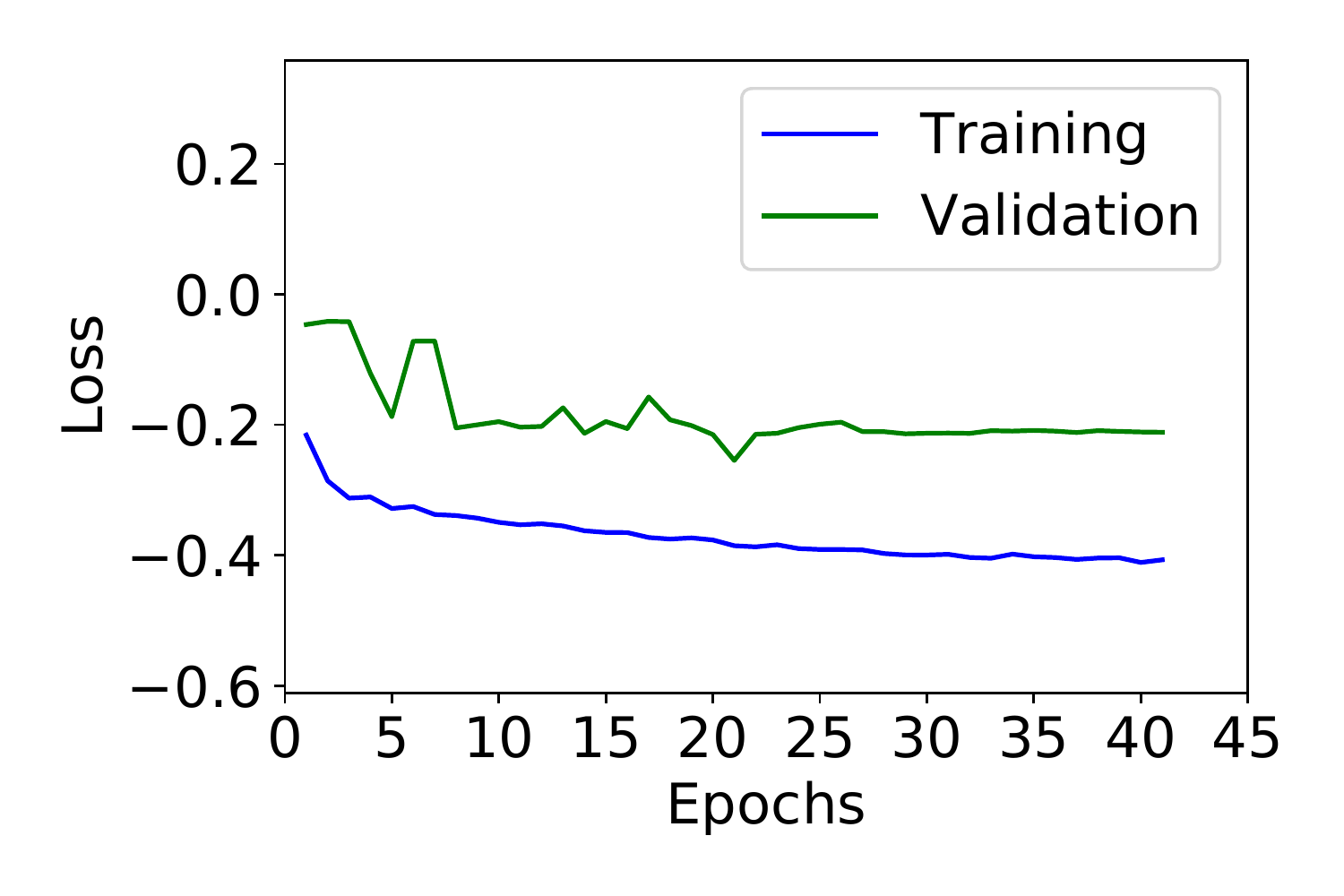}}  
    \subfloat[EM iteration 3]{\includegraphics[width=2.1in]{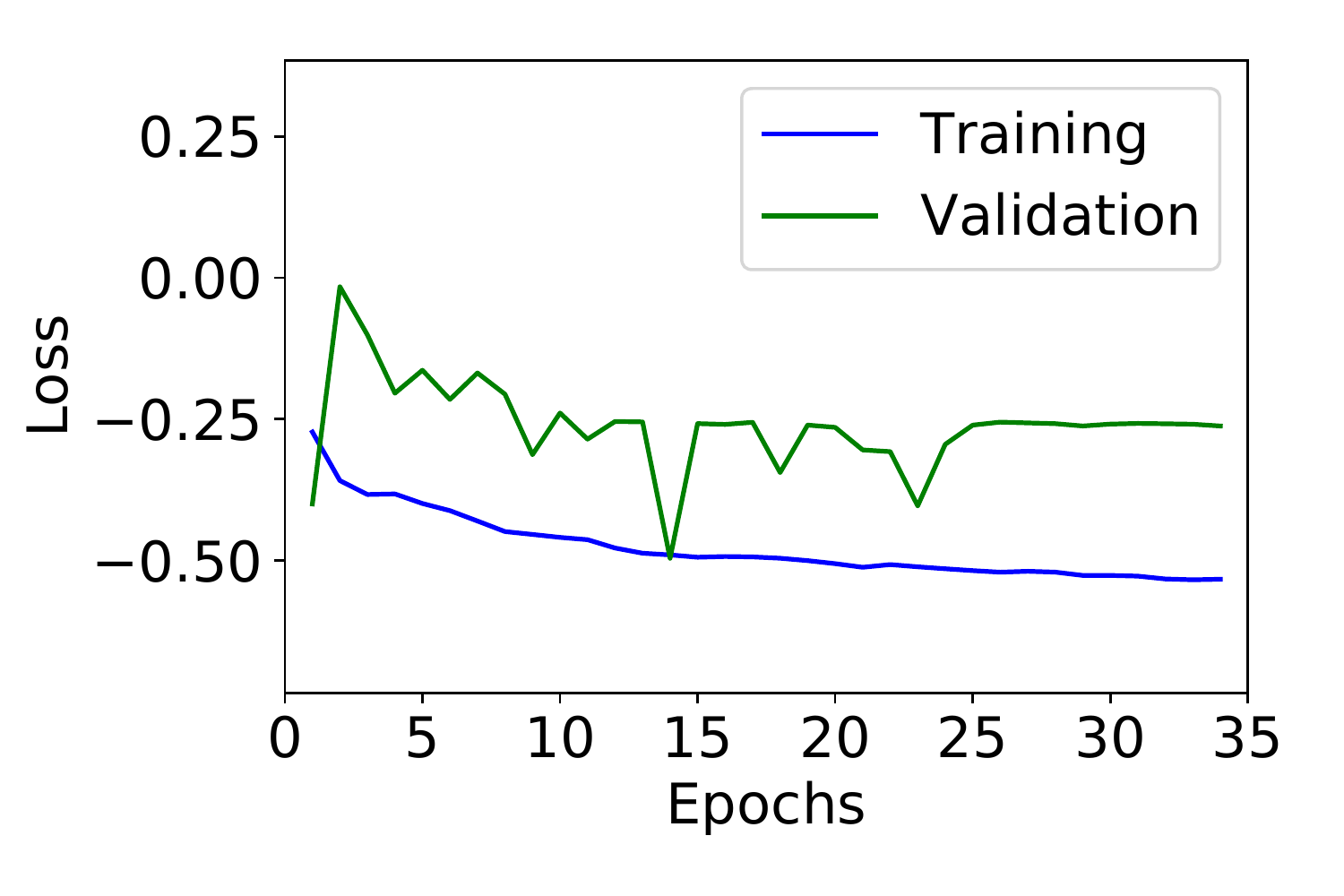}}\\  
    \subfloat[EM iteration 4]{\includegraphics[width=2.1in]{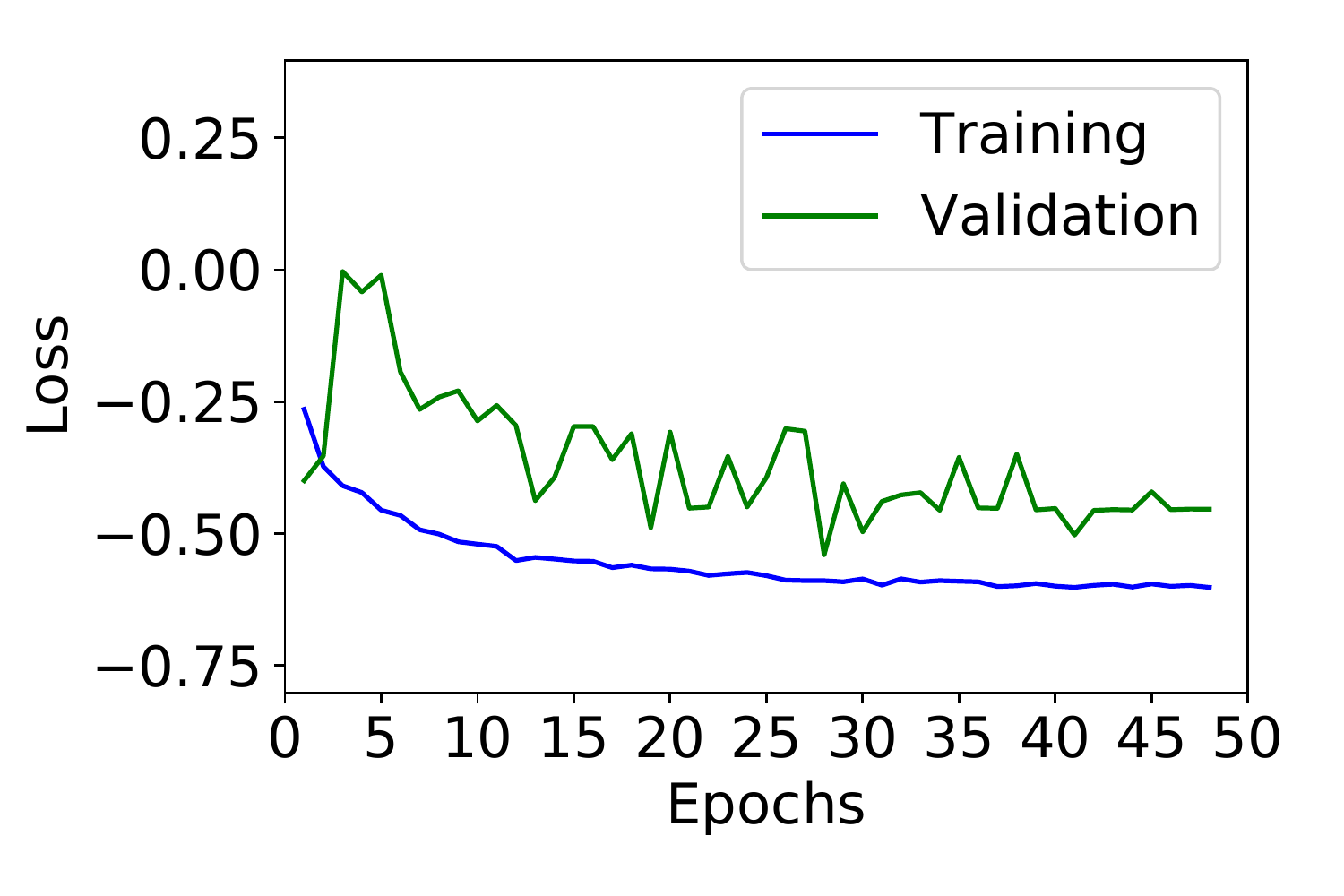}}  
    \subfloat[EM iteration 5]{\includegraphics[width=2.1in]{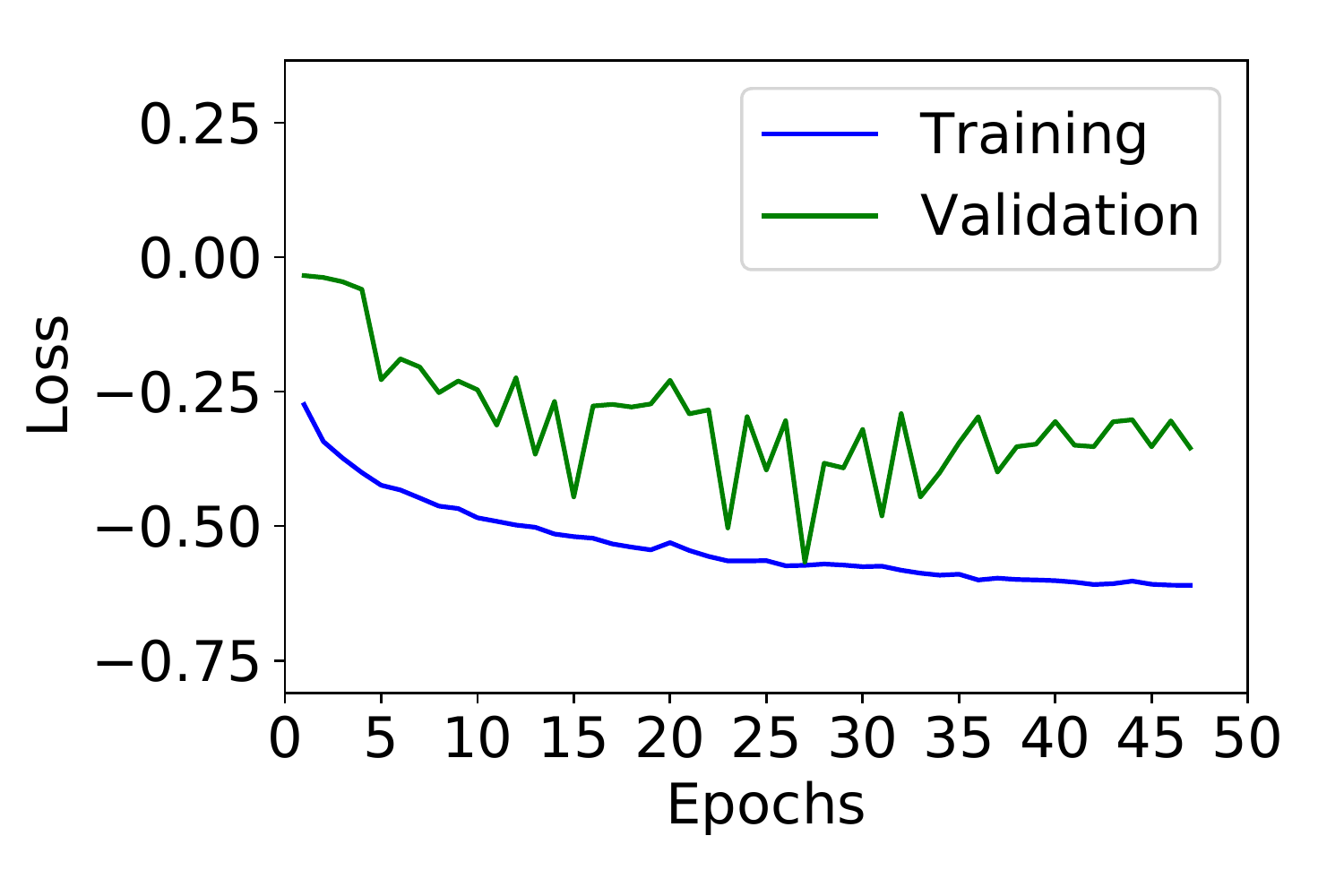}} 
    \subfloat[EM iteration 6]{\includegraphics[width=2.1in]{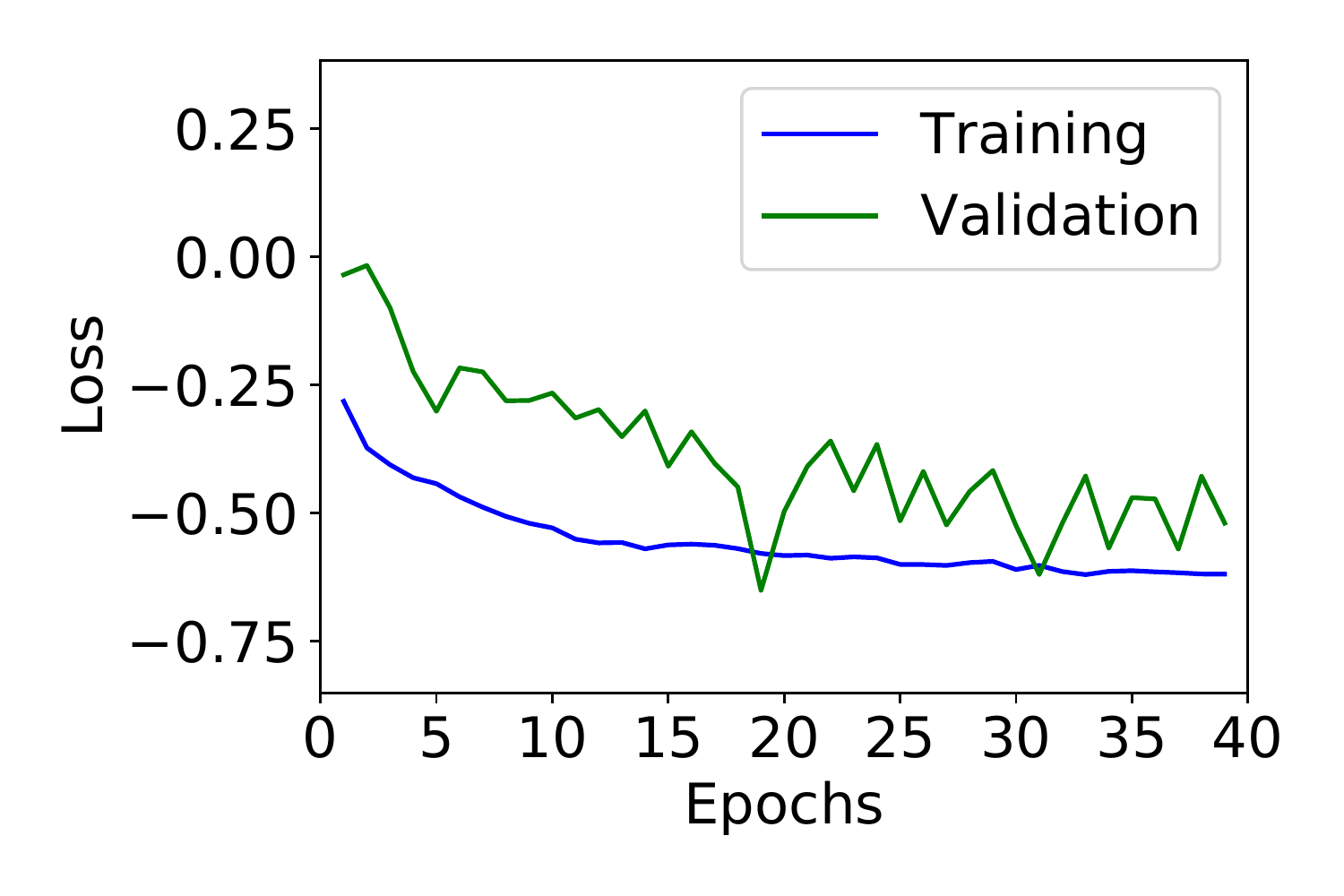}} 
    
    \caption{Training curves of different EM iterations}
    \label{fig:traincurves}
\end{figure*}
\subsection{Comparison on Classification Performance}
We first compared the overall classification performance between the baseline U-Net model and our proposed model. The setup was the same as described at the beginning of this section. The results were summarized in Table~\ref{tab:compare}. The first column in the confusion matrices was the number of pixels predicted into the non-stream class. The second column in the confusion matrices was the number of pixels predicted into the stream class. We can see that the pre-trained U-Net model from imperfect ground truth label had very poor precision and recall in the streamline class (the overall F1 score was 0.46). In contrast, our proposed U-Net with EM iterations improved the precision from 0.39 to 0.66 and improved the recall from 0.57 to 0.58. The overall F1-score in our method is 0.67, significantly higher than the baseline U-Net model. Examine of the confusion matrix shows that our method reduces the number of false positives from 147480 to 48658 (by 67\%) and reduces the number of false negatives from 79854 to 44303 (by 55\%). The metrics confirmed that our proposed method significantly enhanced the baseline image segmentation model when the ground truth segment labels were imperfect. 


\subsection{The learning curve of EM iterations}
\begin{figure}[h]
    \centering
    \includegraphics[width=2.2in]{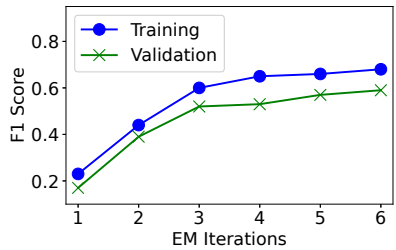}    
    \caption{Training and validation F1-Score after each EM iteration}
    \label{fig:fscorevsiteration}
\end{figure}

In order to understand the influence on the training process of each EM iteration, we plotted the training and validation F1-score (based on the current inferred ``true" label location) after each EM iteration in Figure~\ref{fig:fscorevsiteration}. The F1-score after each EM iteration was from the re-trained U-Net model (through up to 50 epochs) based on the currently inferred label location. We can see that the F1-score continued improving during EM iterations. The training and validation F1-scores in the first EM iteration (slightly above 0.2) were worse than those of the pre-trained U-Net model. The F1-scores significantly improved in the 2nd and 3rd iterations and converged at the sixth iteration (with a validation F1-score of 0.60). 

In order to examine the detailed model learning process, we also plotted the training curves (training and validation loss over every epoch) within each EM iteration. Those training curves were shown in Figure~\ref{fig:traincurves}. As the EM iteration continues, the gap between training loss and validation loss decreased. We also observed that the converged training and validation loss was lower through the EM iterations, largely due to the continual improvement over label locations.



\subsection{Inference of true label locations}
\begin{figure}
    \centering
    \subfloat[Earth imagery background]{\includegraphics[width=1.5in]{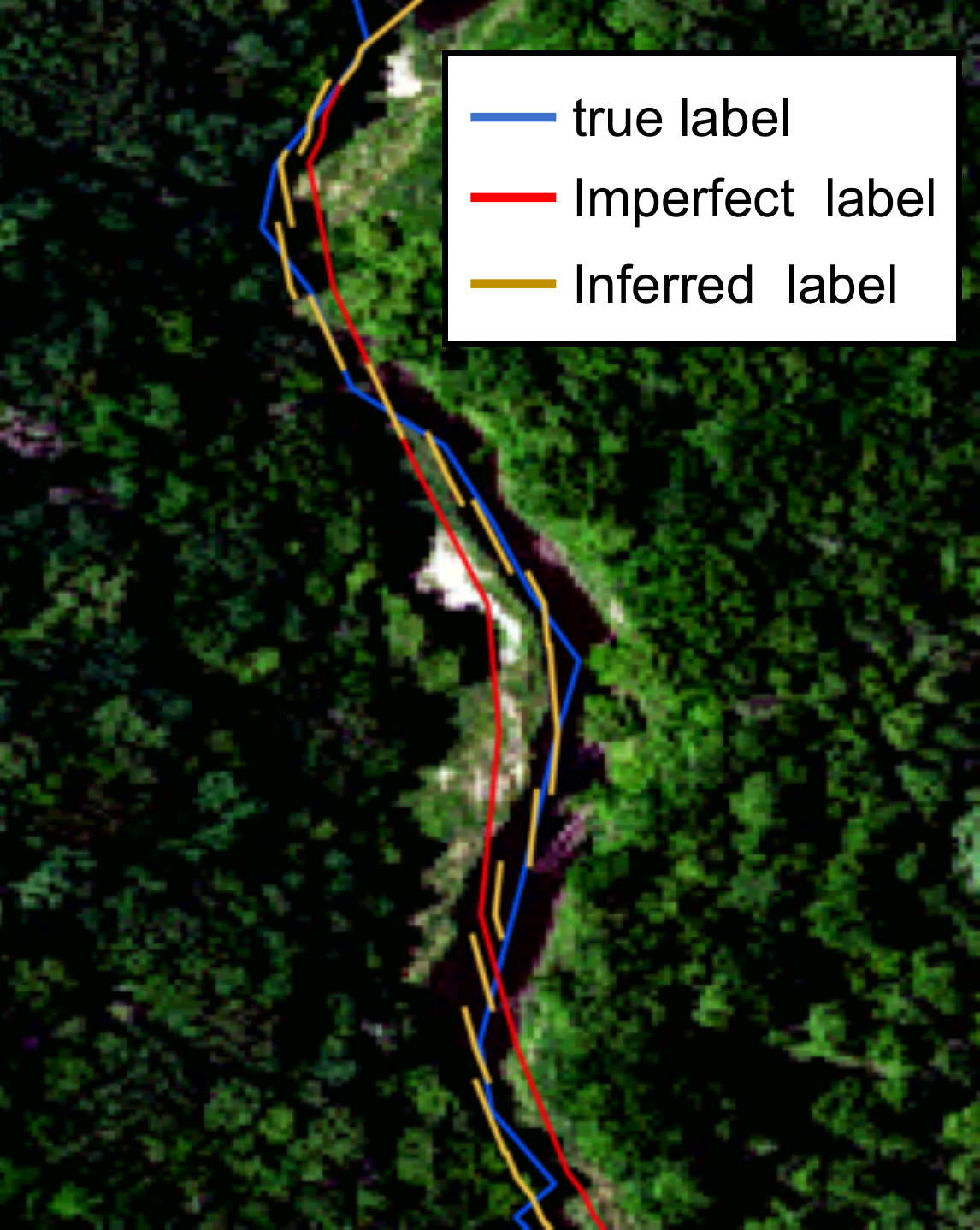}}
    \hspace{2mm}
    \subfloat[River channel background]{\includegraphics[width=1.48in]{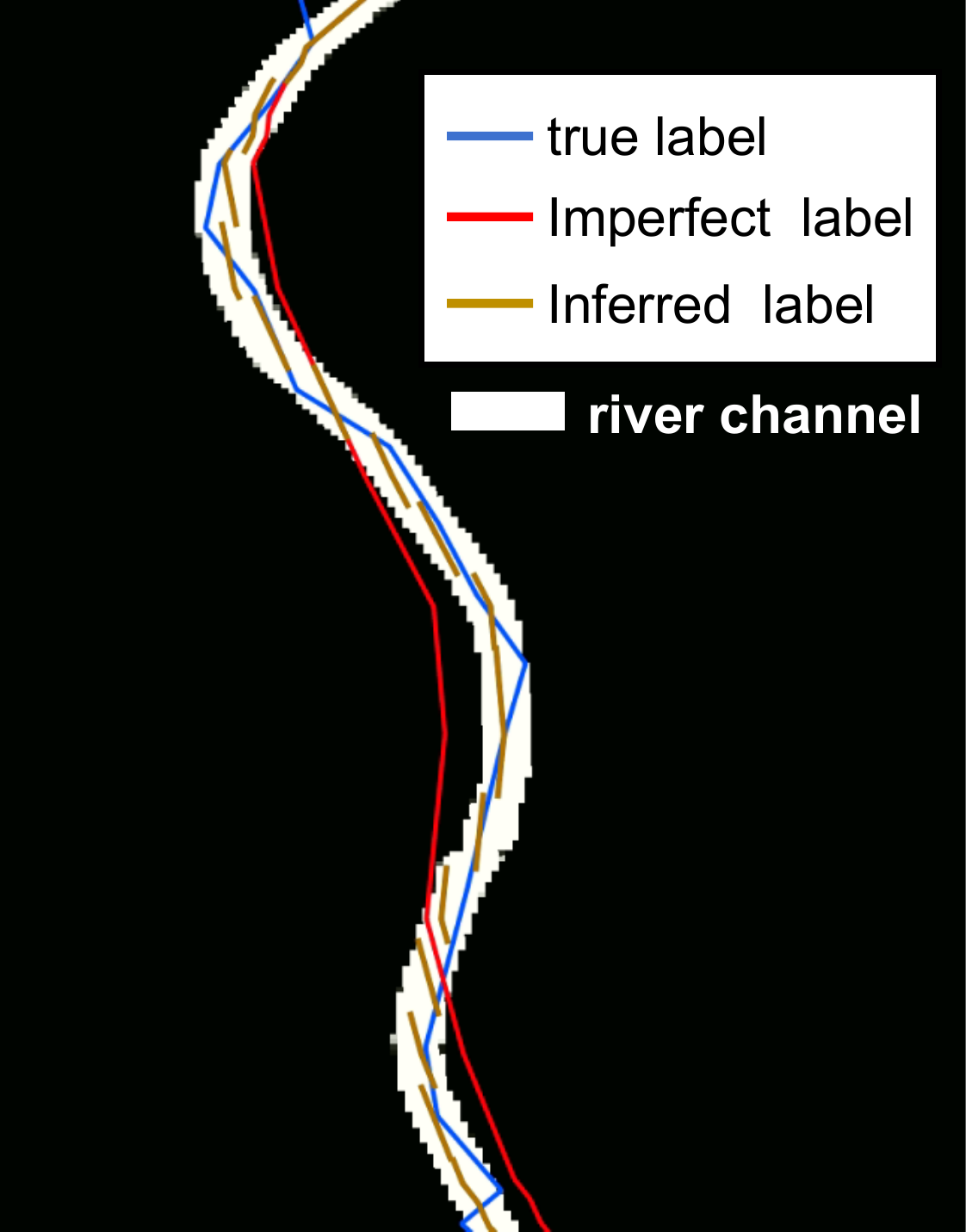}}
    \caption{Visualization of the inferred true label location, manually refined (true) label location, and initial imperfect line location (best viewed in color)}
    \label{fig:candidates}
\end{figure}
\begin{figure*}
    \centering
    \subfloat[Manually refined streamline]{\includegraphics[height=1.3in]{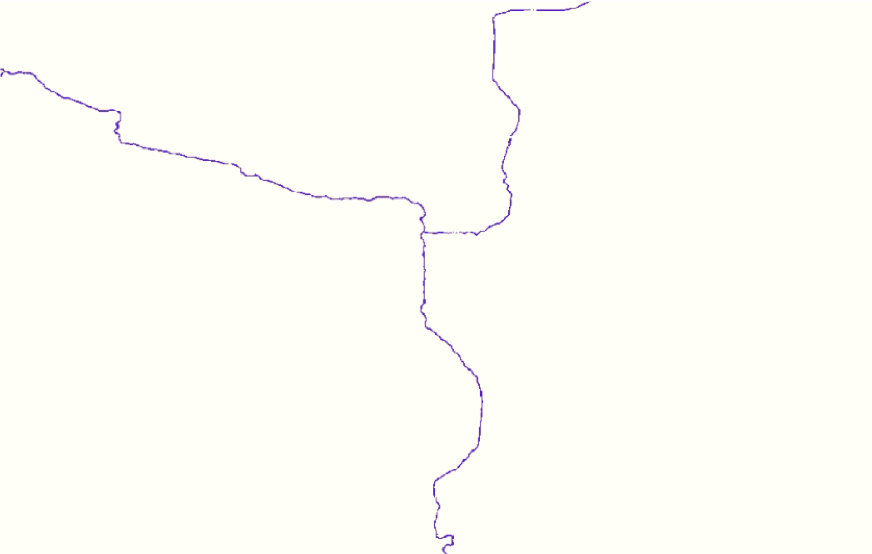}}
    \hspace{2mm}
    \subfloat[U-Net prediction]{\includegraphics[height=1.3in]{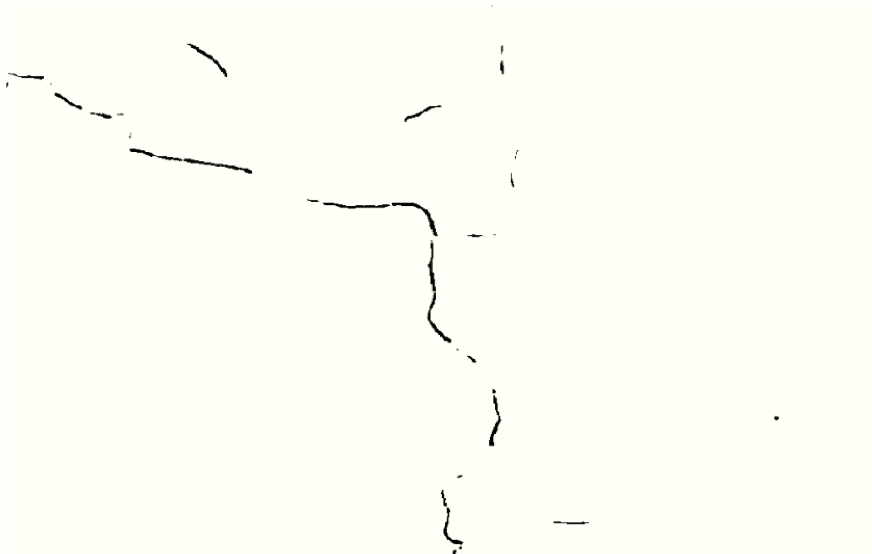}}
    \hspace{2mm}
    \subfloat[Our model prediction]{\includegraphics[height=1.3in]{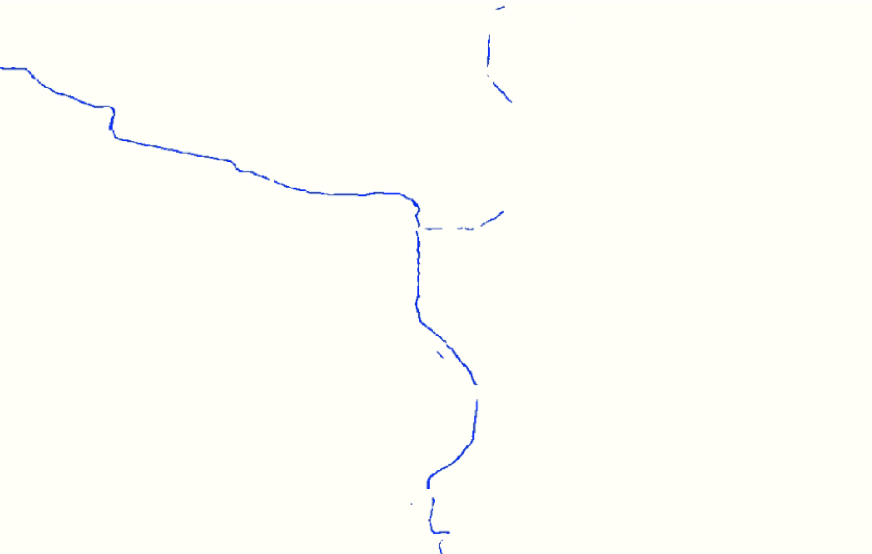}}
    \caption{Visualization of the final predicted class maps in the test region}
    \label{fig:predmaps}
\end{figure*}
We also visualized the inferred (selected) true label locations during the EM iteration. Figure~\ref{fig:candidates} showed the comparison of our inferred true label locations (in brown) with the manually refined true label locations (in blue) as well as the initial imperfect label locations (in red). The actual footprint of the river channel was also shown in the background imagery, including a true color earth imagery in Figure~\ref{fig:candidates}(a) and a binary map in Figure~\ref{fig:candidates}. From the comparison, we can see that our inferred true label locations (those selected candidate segments in brown) were far closer to the manually refined true label locations (in blue) than the initial imperfect label (in red). This visualized results verified that our EM iteration framework could infer the true label locations while training the U-Net model. What was even more interesting was that our algorithm seemed to make the best efforts in inferring true label locations. For example, when the initial imperfect label line segments (in red) were not well oriented with the true line location (in blue), our selected candidates (in brown) still crossed over with the blue line as much as possible. And when the initial imperfect label line had the same orientation as the true line, our inferred line location was almost perfectly aligned with the true line.

\subsection{Interpretation of final prediction map}

We also visualized the predicted streamline class maps in the test region from our model and the U-Net model. Due to limited space, we selected one representative sub-area in the test region to have a zoomed-in view. The results were shown in Figure~\ref{fig:predmaps}. Figure~\ref{fig:predmaps}(a) showed the manually refined streamline labels, which were the ``perfect" ground truth in testing. Figure~\ref{fig:predmaps}(b) showed the predicted streamline locations by the baseline U-Net model. We can see that it contained many false positives (false streamlines predicted) and false negatives (missing true streamlines). For example, the upper right branch of the stream was barely identified by U-Net. The upper left branch was also not continuous in the U-Net predictions. In addition, there were false stream segments in U-Net predictions on the top. In contrast, our results (shown in Figure~\ref{fig:predmaps}(c)) were far better with fewer false positives and false negatives. We also did a careful examine of the entire predicted maps over the entire region and found similar trends as shown in this figure.

\subsection{Analysis of computational time costs}
\begin{figure}[h]
    \centering
    \includegraphics[width=3.1in]{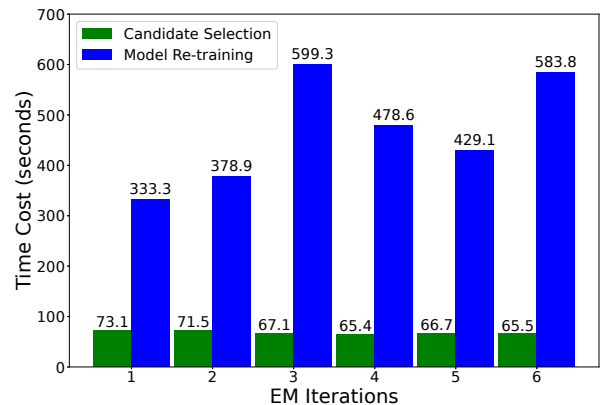}    
    \caption{Time Cost}
    \label{fig:timecost}
\end{figure}

We evaluated the computational efficiency of our proposed EM framework. The experiments were conducted on our deep learning workstation with 4 NVIDIA RTX 6000 GPUs connected by NV-Link (each GPU has 24GB memory). Model training was conducted on all four GPUs through the distributed training tool in Tensorflow. The time costs between EM iterations were summarized in Figure~\ref{fig:timecost}. The green bar and blue bar showed the time costs of candidate selection (re-generating rasterized true label map) in the CPU and model re-training in the GPUs. From the results, we can see that the candidate selection part took far less time than the model training and its time cost was relatively stable across EM iterations. The time cost of model training varied across EM iterations due to early stopping. The longest training time was around 10 minutes in one iteration. The numbers were highly dependent on the hardware platform.
\section{Conclusion and Future Works}
We investigated deep learning models for earth imagery segmentation from imperfect ground truth labels with geometric annotation errors. The problem is important for broad applications in earth science, such as refining the National Hydrologic Dataset through high-resolution remote sensing imagery. However, the problem is non-trivial due to the requirement to infer the true geometric label locations and update neural network parameters simultaneously. We propose a generic deep learning framework based on the EM algorithm to simultaneously infer hidden true label locations and train deep learning model parameters. Evaluations on a real-world hydrological dataset confirmed that the proposed framework significantly outperformed the baseline method.

For future work, we plan to continue to improve the core component in our proposed EM algorithm, i.e., candidate true shape location generation and selection. We also plan to generalize our framework from polyline ground truth shapes to polygon shapes. 

\clearpage
\bibliographystyle{unsrt}
{\small \bibliography{refs}}

\clearpage
\appendix
\section{Implementation Details}




\subsection{U-Net model}

The U-Net model architecture is shown in Table~\ref{tab:unet}. The model consists of five double convolutional layers and max-pooling layers in the downsample path as well as five double convolutional layers and transposed convolutional layers in the upsample path. There is a batch normalization operation within each convolutional layer before non-linear activation based on ReLU (rectified linear unit). The model has 31,455,042 trainable parameters in total. 
\begin{table}[h]\footnotesize
\centering
\caption{U-Net model architecture}
\label{tab:unet}
\begin{tabular}{ccc}
\toprule
Layer (type) & Output shape & Param \# \\ \hline
Input &(None, 224, 224, 4)&0\\ \hline  
Conv2D &(None, 224, 224, 32)&1184\\ \hline  
Conv2D&(None, 224, 224, 32)&9248\\ \hline  
Max pooling&(None, 112, 112, 32)&0\\ \hline 
Conv2D&(None, 112, 112, 64)&18496\\ \hline

Conv2D&(None, 112, 112, 64)&36928\\ \hline  
Max pooling&(None, 56, 56, 64)&0\\ \hline  
Conv2D&(None, 56, 56, 128)&73856\\ \hline  
 Conv2D&(None, 56, 56, 128)&147584\\ \hline
Max pooling&(None, 28, 28, 128)&0\\ \hline
  
 Conv2D&(None, 28, 28, 256)&295168\\ \hline  
 Conv2D&(None, 28, 28, 256)&590080\\ \hline
Max pooling&(None, 14, 14, 256)&0\\ \hline

 Conv2D&(None, 14, 14, 512)&1180160\\ \hline  
 Conv2D&(None, 14, 14, 512)&2359808\\ \hline
Max pooling&(None, 7, 7, 512)&0\\ \hline

 Conv2D&(None, 7, 7, 1024)&4719616\\ \hline  
 Conv2D&(None, 7, 7, 1024)&9438208\\ \hline
Up sampling&(None, 14, 14, 1024)&0\\ \hline
 Conv2D&(None, 14, 14, 512)&7078400\\ \hline  
 Conv2D&(None, 14, 14, 512)&2359808\\ \hline
Up sampling&(None, 28, 28, 512)&0\\ \hline

 Conv2D&(None, 28, 28, 256)&1769728\\ \hline  
 Conv2D&(None, 28, 28, 256)&590080\\ \hline
Up sampling&(None, 56, 56, 256)&0\\ \hline
 Conv2D&(None, 56, 56, 128)&442496\\ \hline  
 Conv2D&(None, 56, 56, 128)&147584\\ \hline
Up sampling&(None, 112, 112, 128)&0\\ \hline

 Conv2D&(None, 112, 112, 64)&110656\\ \hline  
 Conv2D&(None, 112, 112, 64)&36928\\ \hline
Up sampling&(None, 224, 224, 64)&0\\ \hline

Conv2D&(None, 224, 224, 32)&27680\\ \hline  
 Conv2D&(None, 224, 224, 32)&9248\\ \bottomrule

\end{tabular}
\end{table}

 


\subsection{Model training}
For U-Net, we applied the model architecture in Table~\ref{tab:unet} with Keras. In model training, we used Adam optimizer. The loss function was the negative dice coefficient loss. The initial learning rate was $10^{-1}$. We trained the model for 50 epochs with a mini-batch size of $16$. 

\section{Sensitivity Analysis}
We conducted two self-comparisons to test the effect of the hyper-parameters in candidates selection. Specifically, we evaluated the effect of randomness probability $\epsilon$ and maximum candidates number $K$ (We define those parameters in the Evaluation section). When evaluating the effect of randomness probability $\epsilon$, we fixed the maximum candidates number $K= 5$ and increased $\epsilon$ from $0.01$ to $0.2$. Results show that when increasing the candidate random selection chance, the F score at first increases and then decreases. This can be explained by the EM algorithm property. When we increase random selection chance, we can prevent the EM iteration from being stuck in the local minimum. However, if the random selection chance is too large, we have a lower probability to select a good quality candidate. When evaluating the effect of maximum candidates number $K$, we fixed $\epsilon = 0.05$, and increase $K$ from 5 to 19 (All candidates). Results show that the top 5 candidates shows the best F score. When we increase $K$ to 19, the F score drops by $2\% - 4\%$. The results show that too many candidates can decrease the chance of selecting good candidates.
\begin{figure}[h]
    \centering
    \includegraphics[width=3.2in]{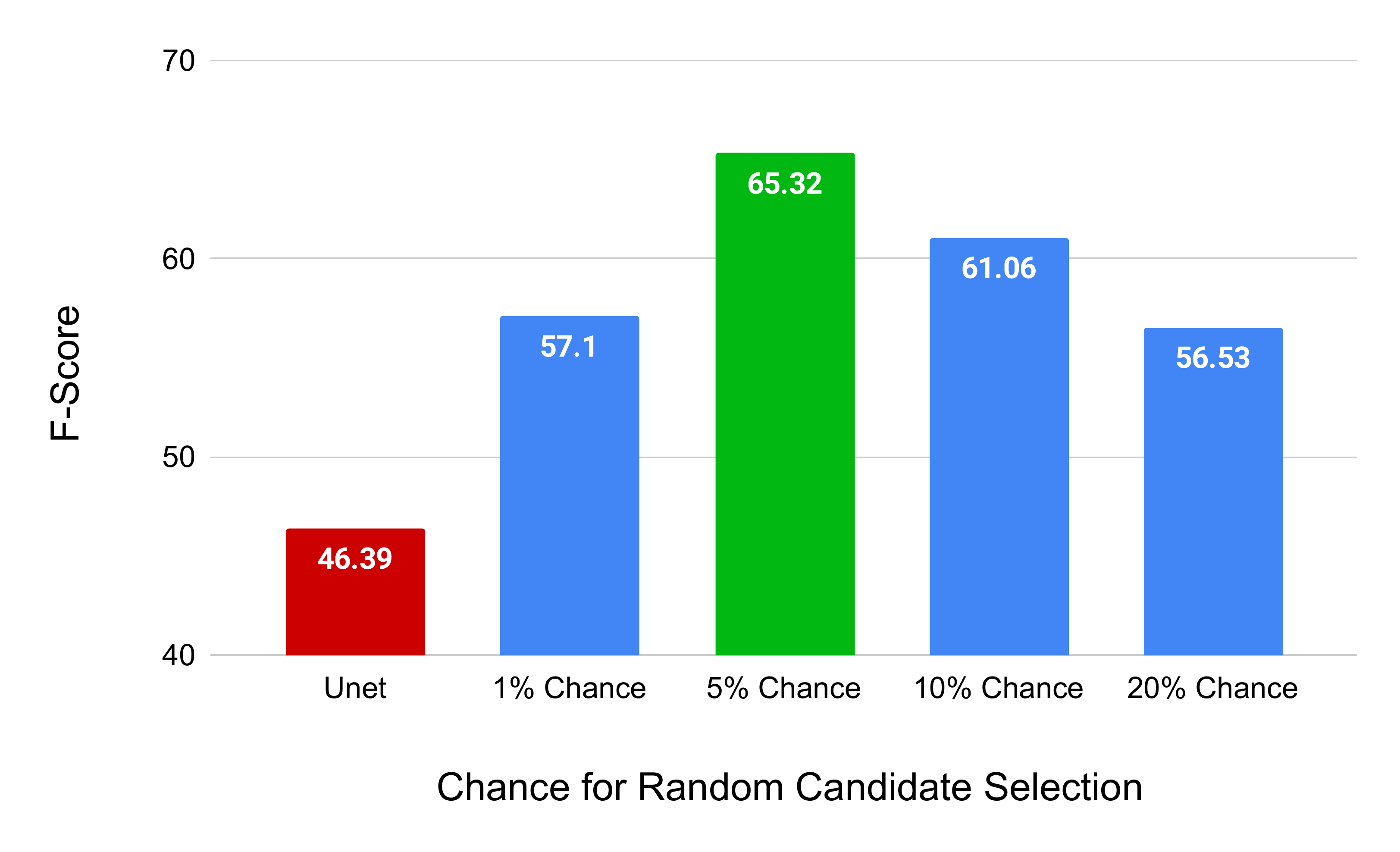}    
    \caption{The effect of random selection chance}
    \label{fig:RandomCandidateChance}
\end{figure}

\begin{figure}[h]
    \centering
    \includegraphics[width=3.2in]{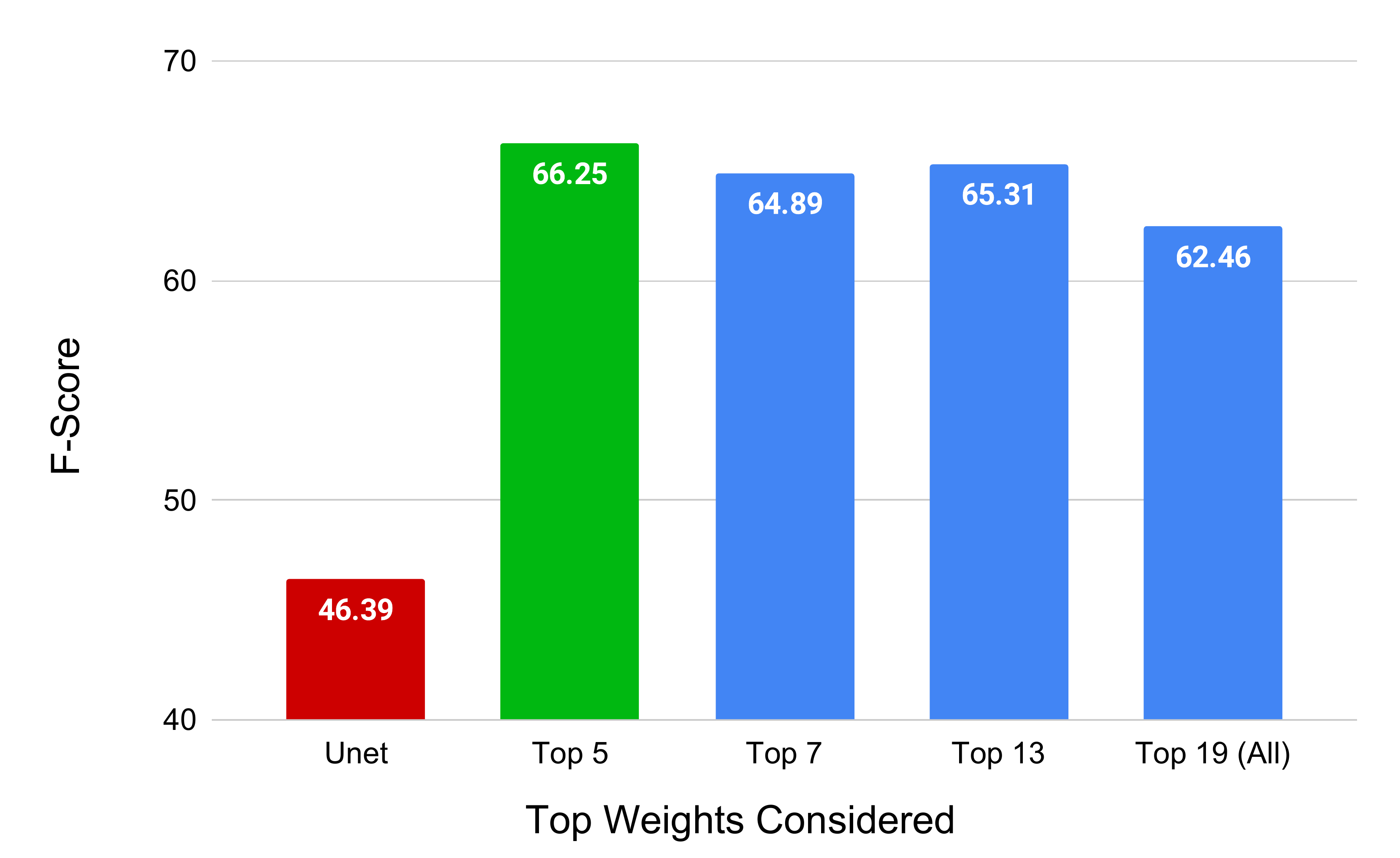}    
    \caption{The effect of top $K$ weights considered}
    \label{fig:TopWeightsConsidered}
\end{figure}

\end{document}